\documentclass[letterpaper, 10 pt, conference]{ieeeconf}
\IEEEoverridecommandlockouts
\usepackage{preamble}

\title{\LARGE \bf
Can Causal Models Enhance Robot Navigation? Online Causal Adaptation for Real-Robot Navigation
}

\author{Zhitao Liang$^{\mathsection}$, Alex Mitrevski, Emmanuel Dean, and Karinne Ramirez-Amaro%
\thanks{This work was supported by the Chalmers Gender Initiative for Excellence (Genie) and partially supported by the Swedish Research Council (Vetenskapsrådet, VR), Project ID: 2025-06377, and the Chalmers AI Research Centre (CHAIR).}
\thanks{The authors are with the Division for Systems and Control, Department of Electrical Engineering, Chalmers University of Technology, Gothenburg, Sweden
        {\tt\small <zhitao, alemitr, deane, karinne>@chalmers.se}}%
\thanks{$^{\mathsection}$Corresponding author}
}

\begin{document}

\maketitle
\thispagestyle{empty}
\pagestyle{empty}

\begin{abstract}
    Causality in robotics aims to produce more interpretable and flexible robot behaviours by enabling robots to predict the consequences of their actions; however, deploying causal models with existing systems (e.g., navigation) operating in real environments remains understudied.
    This paper addresses the challenging problem of transferring causal models in real-robot experiments for a navigation scenario. 
    We study this problem in two ways:
    (i) using the causal model as an offline evaluation module that predicts the competence of recorded real-robot navigation trajectories and relates it to quantitative navigation performance, and
    (ii) using the causal model as an online adaptation module that intervenes when the predicted competence of the default navigation is low.
    We validate our approach in a physical service robot that patrols around corridors.
    We show that the predicted competence correlates positively with path efficiency, and negatively with path irregularities (suboptimal behaviour).
    The model predictions also show strong agreement with human annotations (Cohen’s kappa value of $0.88$). In online experiments, the proposed method improves navigation performance in complex scenarios such as cornering and obstacle avoidance, yielding higher predicted competence and better navigation metrics than the default navigation baseline.
    In simpler scenarios, where the baseline already performs near-optimally, the causal adaptation provides limited benefit.
    These results indicate that causal models are particularly effective in enhancing navigation under increased task complexity.
    Overall, our results demonstrate that causal models developed for behavioural interpretation can be successfully integrated into real-robot navigation systems.
\end{abstract}

    \section{INTRODUCTION}
    \label{sec:introduction}

    Autonomous robot behaviours are typically driven by policies that are either hand-coded, which include implicit expert knowledge about causal structures relevant for producing the appropriate behaviours, or learned from examples, usually acquired using associative learning.
    In both cases, the robot's decision-making process can be difficult to interpret either to outside observers (e.g., when using black-box models) or to the robot itself (e.g., it cannot perform introspection to adapt its behaviour).
    
    An alternative way to produce a robot's behaviour is by using causal representation and reasoning \cite{uhde2020,hellstroem2021,stocking2022,zhu2025,lee2023,castri2025}, where a robot's behaviour is guided by explicit causal models, which are either manually specified or learned.
    Traditionally, causal models have been used to analyze the causes of failures and optionally performing failure correction; this has been done at various levels of the robot decision-making system, such as the level of robot configuration parameters~\cite{fang2026}, robot components~\cite{parker2006}, robot plans~\cite{erdem2012}, individual robot skills~\cite{mitrevski2023}, as well as long-horizon tasks\cite{diehl2023,han2023}.
    Furthermore, recent work has also shown that causal models can be useful to consider in many other domains, including human-centered or multi-agent scenarios~\cite{castri2022,briglia2025,wang2026}, as they can produce behaviours that are both easier to understand (due to the explicit causal structure) and simpler to modify (by performing causal interventions and counterfactual reasoning).
 
    \begin{figure}[t]
        \centering
        \includegraphics[width=\linewidth]{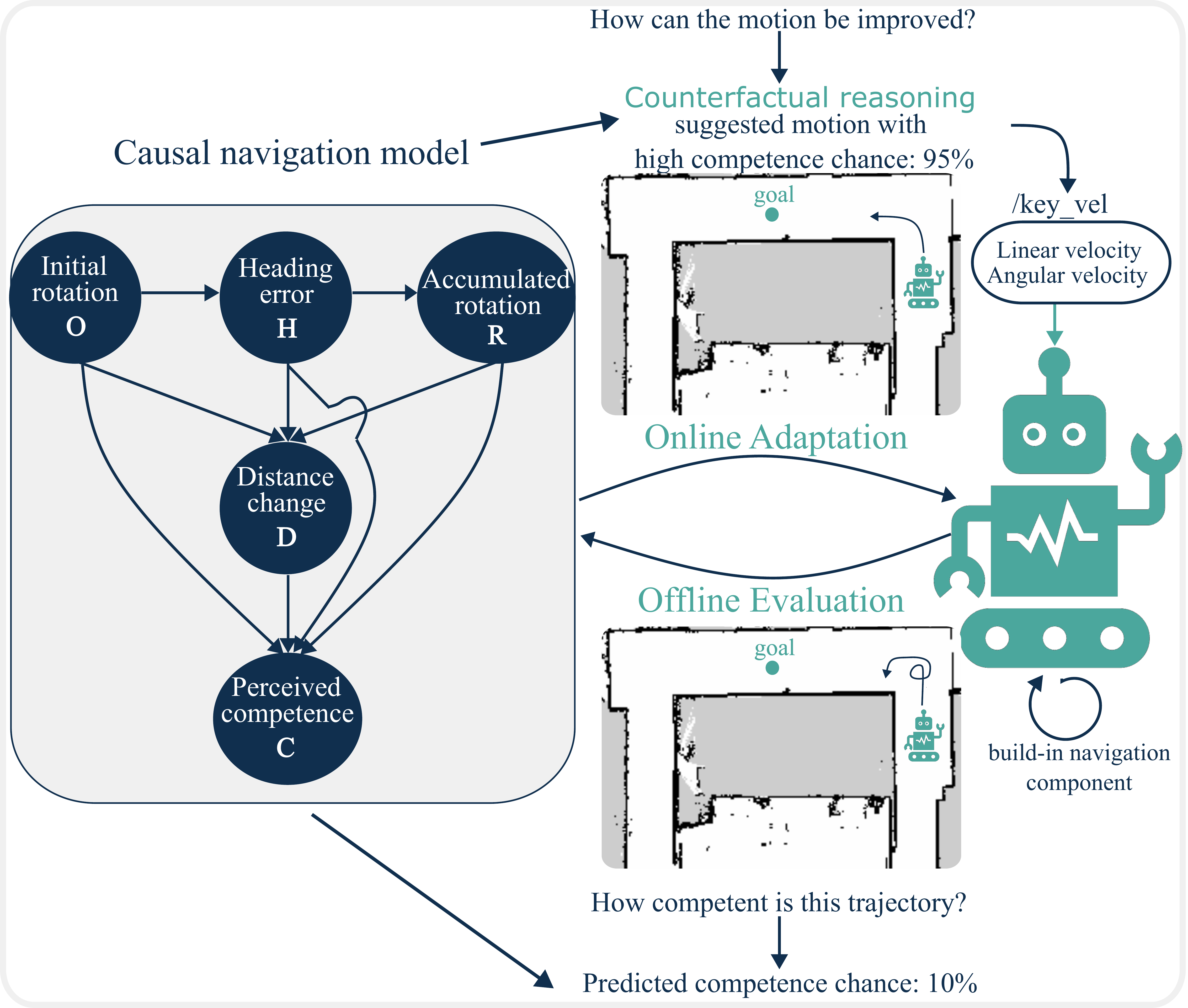}
        \caption{Overview of our method. We use a causal model designed for social navigation both as an offline evaluator that evaluates a human-perceived competence value in robot navigation trajectories and as an online adaptation module that monitors the predicted competence of the robot's default navigation behaviour and intervenes by providing velocity commands when the predicted competence is low during real-world navigation tasks.}
        \label{fig:overview}
        \vspace{-0.5cm}
    \end{figure}
    
    While these developments in robot causality are promising, studies of causal models in human-centered applications usually do not investigate how such models perform under reduced conditions, where some causal variables may not be important to consider.
    In particular, there is a limited understanding of how behaviours guided by causal models compare to traditional robot control approaches that already perform well in many scenarios. This raises two important questions: (i) Can causal-based components provide measurable benefits over standard robot systems?, (ii) Can such models be effectively combined with existing robot control policies rather than replacing them?, and (iii) Can causal models developed for specific navigation contexts be reused in other settings without retraining?

    In this paper, we study these two questions through the lens of a robot's navigation behaviour. We investigate whether a robot can meaningfully benefit from including a causal model in its decision-making to improve the performance of various navigation tasks. 
    An overview of our method is shown in Fig. \ref{fig:overview}.
    We perform a study in which a causal model that has been learned to encode the robot's perceived navigation competence is used as (i) \emph{an offline evaluation module} that assesses trajectories collected during autonomous navigation, and (ii) \emph{an online adaptation module} that queries the model for identifying appropriate actions (ones which would maximise the perceived competence) during online navigation.
    We conduct this study using a PAL TIAGo robot that navigates corridors and employs its default ROS2-based navigation stack (nav2) for autonomous navigation.\footnote{\url{https://docs.pal-robotics.com/25.01/navigation.html}}
    We compare the standard navigation system with an enhanced version in which the causal model monitors and adapts the navigation behaviour.

    The contributions of this paper are:
    (i) a formal framework based on a causal model, both for evaluating offline trajectories and for adapting a robot's motion online,
    (ii) a physical robot study demonstrating that the competence predictions produced by the causal model align both with human annotations of robot behaviour competence and with quantitative navigation metrics, and
    (iii) a real-robot evaluation showing that the enhanced navigation module integrating the causal-based online adaptation improves navigation performance compared to the standard ROS2 navigation stack, as measured by both navigation metrics and predicted competence.\footnote{Accompanying video: \url{https://youtu.be/-aFszuaShOs}}
    Our paper also demonstrates that a causal model that has been learned to maximise positive social perception can be reused in a real-robot navigation setting without retraining, suggesting that the proposed framework increases the reusability and transferability of causal navigation models across deployment contexts.

    \section{RELATED WORK}
    \label{sec:related-work}

    \subsection{Causal Reasoning in Robot Navigation}

    Prior work has increasingly adopted causal inference in robot navigation and autonomous driving to improve the interpretability and robustness of navigation planning policies. An early work employs causal learning to generate counterfactual environment data in vision-and-language navigation tasks~\cite{parvaneh2020}, and others integrate causal structures into learning-based navigation systems, for example, by embedding causal representations within neural architectures such as continuous-time networks and transformers, allowing navigation policy-learning frameworks to reason about causal action-outcome relationships beyond purely correlational learning~\cite{vorbach2021,wang2023,zhang2023,wang2024_neur,wang2024_cvpr,zhao2024,liu2025,wang2025,wang2026}. While these methods demonstrate the value of causal reasoning and counterfactual learning for navigation, causal modules are typically used during learning process or offline decision selection rather than for adapting robot navigation behavior during execution. In particular, most of the work was tested in simulation. Moreover, causal relationships are defined over task-centric variables (e.g., navigation success), and have not been used to explicitly connect human perception or social competence with robot behaviors during navigation.

    \subsection{Causal Reasoning for Behaviour Adaptation}

    Some research has explored causal reasoning as an approach for robot behavior adaptation in response to failures or uncertainty in robot manipulation tasks. For example, Diehl et al.~\cite{diehl2023} employed a Causal Bayesian Network to infer the relevant features that caused robot failures, therefore explaining the observed failures, predicting task outcome, and guiding corrective actions in the robotic system. Another work used causal probabilistic reasoning frameworks to support behavior adaptation under uncertainty, for instance, by adjusting system configurations or selecting alternative actions based on the inferred causal relationships~\cite{cannizzaro2025}.

    \subsection{Causal Reasoning in Social Contexts}

    Causal reasoning has also been investigated in social and human-robot interaction contexts to understand and model the dynamics of human behavior and interaction with robots. Castri et al.~\cite{castri2022} modeled causal relationships among human motion variables, environmental factors, and interaction outcomes to discover the dynamics in human spatial interactions. Some other work investigated how robot actions may causally affect human reactions and responses~\cite{tung2024} and the use of large-scale simulated environments to extend causal learning in complex social settings~\cite{mcduff2022}. These approaches focus on understanding human-robot interaction dynamics in simulated settings rather than explicitly modeling how robot navigation behavior causally affects human perceived navigation competence and investigating if the causation potentially improves robot behaviors during execution. 

    \subsection{Human Perception of Robot Navigation}

    Human perception has been widely explored as a critical factor in the evaluation of socially aware robot navigation. Several works proposed human studies to collect ratings that reflect how competent, clear, or appropriate a robot's navigation behavior is perceived by observers, typically using Likert-scale questionnaires~\cite{gao2022evaluation,mavrogiannis2022social}. Datasets and benchmarking frameworks have been proposed to support systematic collection of human perception data in navigation scenarios in simulated settings. SEAN together dataset~\cite{tsoi2022sean,zhang2025predicting,zhang2025few} collected a structured dataset in which human participants evaluated a dynamic HRI task, ``Robot-Following task", where a robot guides a person to a pre-specified goal in a VR simulation of a warehouse, and proposed that perceived competence can be inferred and predicted from navigation trajectories and contextual features. However, in existing work, human perceived competence mainly serves as an evaluation signal for model prediction, but is not used to guide online adaptation of navigation behaviors.

    Tab. \ref{tab:state_of_the_art_summary} summarises the comparison of our work with the related work.
    \begin{table}[t]
        \scriptsize
        \centering
        \caption{Comparison of our framework to the related work}
        \label{tab:state_of_the_art_summary}
        \begin{tabular}{|M{0.1\linewidth}|M{0.15\linewidth}|M{0.175\linewidth}|M{0.15\linewidth}|M{0.15\linewidth}|}
            \hline
            \cellcolor{gray!10!white}\textbf{Ref.} &
            \cellcolor{gray!10!white}\textbf{Causality-based navigation} &
            \cellcolor{gray!10!white}\textbf{Causal model used for behaviour adaptation} &
            \cellcolor{gray!10!white}\textbf{Real-world testing} &
            \cellcolor{gray!10!white}\textbf{Human perception modeled}
            \\ \hline
            ~\cite{castri2025}     & \cmark & \cmark & \xmark & \xmark \\ \hline
            ~\cite{diehl2023}      & \xmark & \cmark & (\cmark) & \xmark \\\hline
            ~\cite{castri2022}     & \xmark & \xmark & \cmark & \xmark \\ \hline
            ~\cite{briglia2025}        & \xmark & offline& \xmark & \xmark \\ \hline
            ~\cite{wang2026}        & \cmark & offline & \cmark & \xmark \\ \hline
            ~\cite{vorbach2021}    & \cmark & offline & \xmark & \xmark \\ \hline
            ~\cite{wang2023}       & \cmark & offline & \xmark & \xmark \\ \hline
            ~\cite{zhang2023}       & \cmark & offline & \xmark & \xmark \\ \hline
            ~\cite{wang2024_neur}  & \cmark & offline & \xmark & \xmark \\ \hline
            ~\cite{wang2024_cvpr}  & \cmark & offline & \xmark & \xmark \\ \hline
            ~\cite{zhao2024}       & \cmark & offline & \xmark & \xmark \\ \hline
            ~\cite{liu2025}        & \cmark & offline & \xmark & \xmark \\ \hline
            ~\cite{wang2025}        & \cmark & offline & \cmark & \xmark \\ \hline
            ~\cite{cannizzaro2025} & \xmark & \cmark & (\cmark) & \xmark \\ \hline
            ~\cite{mcduff2022}     & \cmark & \xmark & \xmark & \xmark \\ \hline
            ~\cite{tung2024}       & \cmark & \xmark & \xmark  & \xmark \\ \hline
            ~\cite{hossen2023}     & \cmark & \cmark & \cmark & \xmark \\\hline
            ~\textbf{Ours}         & \cmark & \cmark & \cmark & \cmark  \\\hline
        \end{tabular}
        \vspace{-0.3cm}
    \end{table}

    \section{BACKGROUND}
    \label{sec:background}

    \subsection{Causal Bayesian Networks}

    A causal model \cite{pearl1994} represents knowledge about a particular domain using a Causal Bayesian Network (CBN), which is defined as a Directed Acyclic Graph (DAG) $\mathcal{G}=(\mathbf{X}, E)$~\cite{JSSv035i03}.
    In the context of robot task execution, the nodes $\mathbf{X}$ denote $N$ variables describing the robot and environment states, e.g., robot behavior variables and human perception outcomes, while the edges $E$ define the causal relations between the variables.
    The DAG and Markov property of Bayesian networks factorise the joint probability distribution of $\mathbf{X}$ into a set of local probability distributions, where each random variable $X_i$ depends on its parent variables $\Pi_{X_i}$:
    \begin{equation}
        P(X_1, X_2, ..., X_N) = \prod_{i=1}^N P(X_i|\Pi_{X_i})
        \label{eq:bayes_net_factorization}
    \end{equation}
    In some cases, the Bayesian networks are manually specified~\cite{cannizzaro2025, diehl2026causalapproachpredictingimproving}.

    \subsection{Causal Navigation Model}
    \label{sec:background-causalNavigation}

    In this work, we utilise a causal model $\mathcal{M}$ specifically designed for social navigation~\cite{diehl2026causalapproachpredictingimproving, zhang2025predicting}.
    In this model, robot navigation behavior is encoded using seven discrete variables that summarise motion over a fixed temporal window; five of these are observation variables and two model a human's perception of the robot's behaviour.
    Concretely, the observation variables in $\mathcal{M}$ are: robot-goal distance change $\mathbf{D}$ (meter), heading error change of the robot orientation with respect to the goal $\mathbf{H}$ (rad), accumulated rotation $\mathbf{R}$ (rad) over a time window, initial orientation $\mathbf{O}$ (rad), and human pose change $\Delta$ (distance from human to robot: meter).
    On the other hand, the human perception variables model the human-perceived robot intention \textbf{I}, and the robot competence $C$, both of which are represented as categorical nodes that are grounded in human rating data collected through user studies~\cite{zhang2025predicting}.
    Given an observed navigation episode, the robot motion within each temporal window is sampled into a fixed-length trajectory representation that is used to instantiate the observational nodes $\mathbf{x}$, and then $\mathcal{M}$ can be used to predict $C$ as $P_{\mathcal{M}}(C =1|\Pi_{C}=\mathbf{x})$ (for simplicity, we denoted $P_{\mathcal{M}}(C =1)$ as $P_{\mathcal{M}}(C)$ throughout the rest of the paper).
    To learn the model, trajectory data (given as a time series) are discretised by K-means clustering, with each cluster centroid representing a characteristic motion pattern, as further explained in \cite{diehl2026causalapproachpredictingimproving}.
    The model is trained with data collected in a virtual environment with a simulated robot.

    The original model was designed for human-aware navigation scenarios involving interactions between the robot and nearby humans. Our objective, however, is to investigate whether a causal model learned in such complex social navigation settings can still provide meaningful guidance in simpler navigation tasks that do not explicitly involve human interaction. In this work, we use a simplified version of the model introduced in \cite{diehl2026causalapproachpredictingimproving}, in which the variables $\Delta$ and \textbf{I} are omitted, because, in our study, we did not include human features in our navigation scenarios. 
    It should be noted that $\Delta$ and \textbf{I} are child nodes, and thus their removal has no effect on the conditional probability tables of the remaining nodes.

    \section{METHODOLOGY}
    \label{sec:methodology}

\begin{figure}[t]
        \centering
        \includegraphics[width=0.6\linewidth]{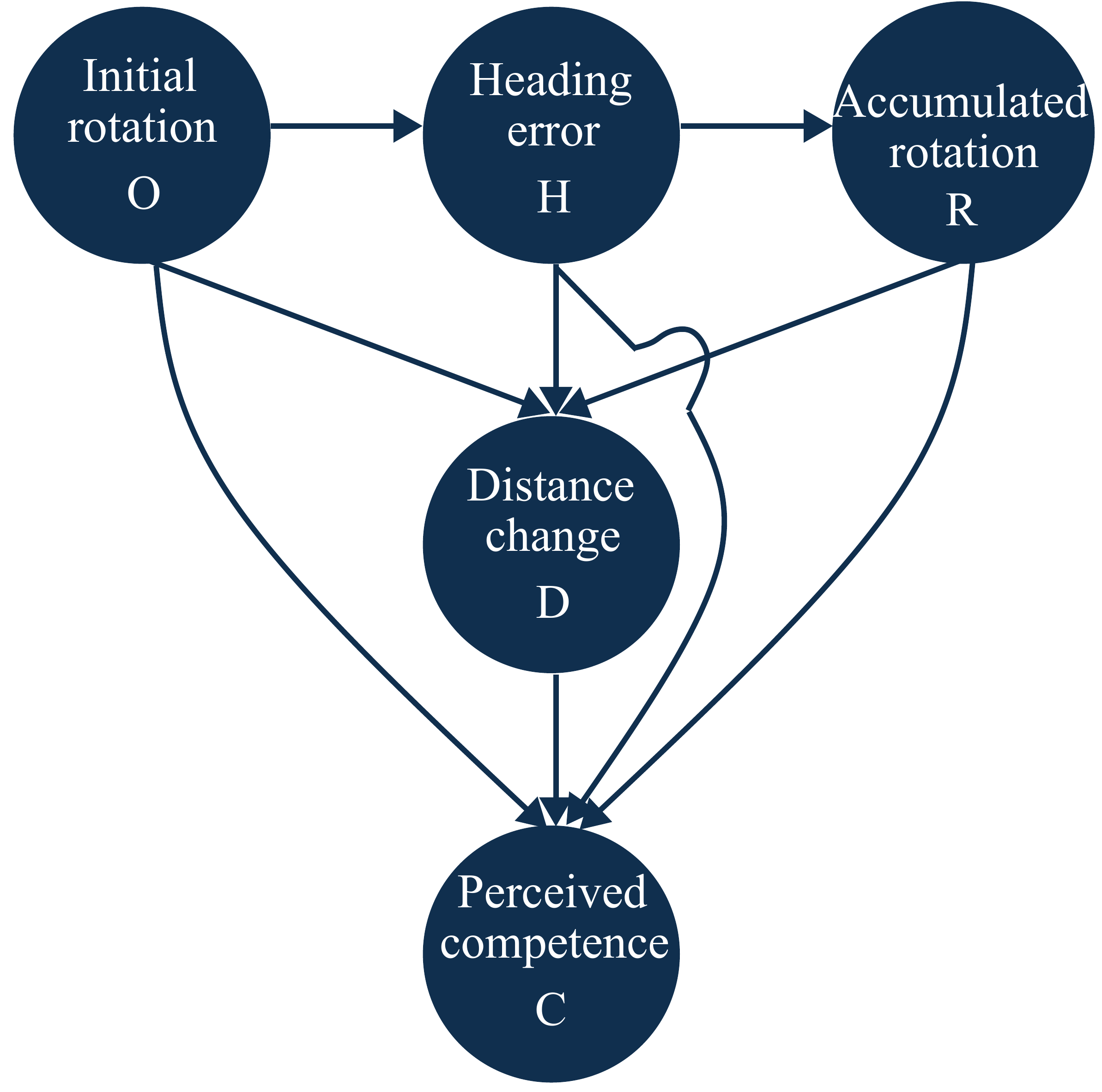}
        \caption{Causal navigation model used in this work (a reduced version of the model described in Sec.~\ref{sec:background-causalNavigation})}
        \label{fig:causal-navigation-model}
        \vspace{-0.3cm}
    \end{figure}

    The objective of this work is to investigate (i) whether a causal model originally developed for social navigation can be reused in real-world navigation scenarios and (ii) whether integrating this model with an existing navigation stack can enhance the robot’s navigation behaviour.
    In particular, we examine whether the competence variable encoded in the simplified causal navigation model obtained from simulations~\cite{diehl2026causalapproachpredictingimproving}, depicted in Fig.~\ref{fig:causal-navigation-model}, can provide a meaningful signal for evaluating and improving real-robot navigation behaviour, even in scenarios where no human interaction is present.
    To study those two questions, we propose using a causal navigation model in two distinct modes:
    (i) an offline evaluation module that evaluates collected navigation trajectories, and
    (ii) an online adaptation module that uses the causal model to direct the robot's navigation during online execution.
    We formalise these two separate modes below.

    \subsection{Causality-Based Offline Evaluation Module}
    \label{sec:methodology-critic}

    Let $\mathcal{T}_i$, $1 \le i \le n$, denote the trajectories generated by the robot while executing $n$ navigation tasks. Each trajectory $\mathcal{T}_i$ records a sequence of robot states during navigation.
    As described in Sec.~\ref{sec:background-causalNavigation}, we divide each trajectory $\mathcal{T}_i$ into temporal windows of $t$ seconds. 
    The $k$-th window of trajectory $\mathcal{T}_i$ is denoted as $\mathcal{T}_i^{(k)}$.
    Within each window, the robot motion is uniformly sampled into $\eta$ states. The $j$-th sampled state of window $\mathcal{T}_i^{(k)}$ is denoted as $\mathbf{s}_{i,j}^{(k)} = (\mathbf{p}_{i,j}^{(k)}, \mathbf{v}_{i,j}^{(k)})$, where $\mathbf{p}_{i,j}^{(k)} = (x_{i,j}^{(k)}, y_{i,j}^{(k)}, \theta_{i,j}^{(k)})$ represents the robot's pose in a global coordinate frame and $\mathbf{v}_{i,j}^{(k)} = (\dot{x}_{i,j}^{(k)}, \dot{y}_{i,j}^{(k)}, \omega_{i,j}^{(k)})$ denotes the velocity expressed in the robot's local coordinate frame. In this work, we use a window duration $t=8\, s, \, \text{and } \eta = 40$ samples per window. We adopt the same temporal resolution and sampling strategy used in the causal model described in Sec.~\ref{sec:background-causalNavigation} to remain consistent with the model design and its training data. This provides a compact representation of the robot motion while preserving the temporal structure needed to estimate navigation behaviour features.
    
    The use of the causal model as an offline evaluation module focuses on this competence evaluation of collected navigation trajectories.
    To verify whether the competence variable predicted by the model aligns with robot navigation performance, we
    (i) leverage numerical trajectory metrics to describe and quantify the robot motion behaviors, and
    (ii) analyse the correlation between model-predicted human-perceived competence and the behavior metrics.
    
    \paragraph{Social Navigation Metrics} Each of the trajectory segments $\mathcal{T}_i^{(k)}$ is evaluated using the following social navigation metrics capturing orientation behaviour, stop frequency, progress towards the goal, and path efficiency:

\noindent\textbf{Path Irregularity ($PI$)}~\cite{gao2022evaluation, biswas2022socnavbench} calculates the accumulated excess rotation beyond the minimum required orientation change, normalised by path length, thereby summarising trajectory smoothness and penalising unnecessary turning:
        \begin{equation}
            PI = \frac{\mathcal{R}_{\mathcal{T}_i^{(k)}} - \text{Min}_\mathcal{R}}{\mathcal{L}_{\mathcal{T}_i^{(k)}}}
        \end{equation}
        Here, the path length $\mathcal{L}_{\mathcal{T}_i^{(k)}}$ of $\mathcal{T}_i^{(k)}$ is accumulated by the length over $\eta$ steps
        \begin{equation}
            \mathcal{L}_{\mathcal{T}_i^{(k)}} = \sum_{j=1}^{\eta}||\vec{p}_{i,j+1}^{(k)} - \vec{p}_{i,j}^{(k)}||_2
        \end{equation}
        The excess rotation $\mathcal{R}_{\mathcal{T}_i^{(k)}}$ is accumulated by the absolute angle the robot has rotated in each step:
        \begin{equation}
            \mathcal{R}_{\mathcal{T}_i^{(k)}} = \sum_{j=1}^{\eta}||\theta_{i,j+1}^{(k)} - \theta_{i,j}^{(k)}||_2
        \end{equation}
        The minimum required orientation change $\text{Min}_\mathcal{R}$ is defined as the initial heading error between the robot’s orientation and the direction towards the goal (i.e., how much the robot needs to rotate to face the goal).
        
\smallskip
\noindent\textbf{Stop Ratio ($SR$)} is defined as the fraction of time steps in which the robot stays approximately static (i.e., step length $||\vec{p}_{i,j+1}^{(k)} - \vec{p}_{i,j}^{(k)}||_2 < \delta$), where $\delta$ is a threshold that depends on the sensor noise. $SR$ captures haulting and hesitation behavior of the robot:
        \begin{equation}
            SR = \frac{\# \left(||\vec{p}_{i,j+1}^{(k)} - \vec{p}_{i,j}^{(k)}||_2 < \delta\right)}{\eta} \quad \text{for } 1\leq j \leq \eta
        \end{equation}

\noindent\textbf{Goal-directed Progress ($GP$)} is defined as the fraction of time steps where the distance $d$ to the goal decreases, and thus evaluates the progress towards the goal:
        \begin{equation}
            GP = \frac{\# \left( (d_{j+1} - d_j) <0 \right)}{\eta} \quad \text{for } 1\leq j \leq \eta
        \end{equation}

\noindent\textbf{Path efficiency ($PE$)}~\cite{schneider2005discussion} is the ratio between the straight-line distance from the initial robot position $\vec{p}_{i,1}^{(k)}$ and the final position $\vec{p}_{i,\eta}^{(k)}$ and the actual path length $\mathcal{L}_{\mathcal{T}_i^{(k)}}$: 
        \begin{equation}
            PE = \frac{||\vec{p}_{i,\eta}^{(k)} - \vec{p}_{i,1}^{(k)}||_2}{\mathcal{L}_{\mathcal{T}_i^{(k)}}}
        \end{equation}

    \paragraph{Human-Model Competence Agreement} Furthermore, given $\mathcal{M}$, we compute the competence $C$ for each of the trajectory segments $\mathcal{T}_i^{(k)}$; this results in a set $\tilde{L}$ of binary labels, where
    \begin{equation}
        \tilde{L}_{i}^{(k)} = \twopartdef{1}{P_{\mathcal{M}}(C|\mathcal{T}_i^{(k)})> \tau}{0}{\text{otherwise}}
    \end{equation}
    with a predefined threshold $\tau$. $\tilde{L}$ provides information on the model's \emph{believed} competence of the robot; however, this may not necessarily coincide with actual navigation quality or with the competencies that would be perceived by a human observer.
    To measure the quality of $\mathcal{M}$ as a navigation evaluation module, we also manually annotate a subset of the trajectory segments, and produce a set $L$ of labels that represent the perceived competence of the robot from an operator's perspective.
    Given $L$ and $\tilde{L}$, we then compute the alignment between the human annotations and the model's believed competences; we use Cohen's kappa coefficient \cite{cohen1960} for this purpose:
    \begin{equation}
        \kappa_{L,\tilde{L}} = \frac{P_o - P_c}{1 - P_c}
    \end{equation}
    where $P_o$ is the proportion of competence labels where $L$ and $\tilde{L}$ are the same, and $P_c$ is the proportion of labels where agreement is expected by chance.
    
    The insight about the competence in prior trajectories could be used to communicate the robot's belief about its own performance to users, but could also guide the robot's learning process so that it can improve its navigation behaviour.
    Either way, the quality of $\mathcal{M}$ as an offline evaluator, expressed through $\kappa_{L,\tilde{L}}$, provides information on \emph{whether} the believed competences can be trusted for those purposes.

    \subsection{Causality-Based Online Adaptation Module}
    \label{sec:methodology-actor}

    The use of $\mathcal{M}$ as an offline evaluator evaluates already executed trajectories, but does not guide a robot's behaviour during online trajectory execution.
    The causal model can, however, be readily incorporated into a continuous control loop, with an objective of finding robot actions that would lead to a high perceived robot competence.
    We achieve this by integrating the model with a given robot navigation component $\mathcal{N}$.
    The idea is that $\mathcal{N}$ handles the robot's navigation behaviour in general, but $\mathcal{M}$ modifies the behaviour whenever an improvement of the perceived competence is considered necessary.
    
    Let the robot's state at time $t$ be defined by the robot's pose $\vec{p}_t$, expressed in a global coordinate frame as in Sec. \ref{sec:methodology-critic}, and the action $\vec{a}_t$ be given by the robot's velocity $\vec{v}_t$, expressed in the robot's local coordinate frame.
    The objective of the online adaptation is to find an action $\vec{a}^*_t$ for which
    \begin{equation}
        P_{\mathcal{M}}(C|\vec{s}_t,\vec{a}^*_t) > \tau
        \label{eq:actorObjective}
    \end{equation}

    Let $\tilde{\vec{a}}_t$ be the action selected by $\mathcal{N}$, such that the believed competence of $\tilde{\vec{a}}_t$ is $P_{\mathcal{M}}(C|\vec{s}_t,\operatorname{do}(\tilde{\vec{a}}_t))$, namely the $\operatorname{do}$-operator \cite{pearl1994} is used to evaluate the competence in the hypothetical case where $\vec{a}_t$ is used as an action.
    In the case where $P_{\mathcal{M}}(C|\vec{s}_t,\operatorname{do}(\tilde{\vec{a}}_t)) < \tau$, we need to find an alternative action $\vec{a}^*_t \neq \tilde{\vec{a}}_t$ that would satisfy Eq. \ref{eq:actorObjective}.
    The identification of $\vec{a}^*_t$ is performed using counterfactual reasoning on $\mathcal{M}$, similar to \cite{diehl2023}.
    Formally, a contrastive search process is performed that attempts to identify the smallest possible change in $\tilde{\vec{a}}_t$ that leads to the desired outcome.
    Let us denote the modified actions that are attempted during the process by $\vec{a}'_t$, and their evaluations under $\mathcal{M}$ as $P_{\mathcal{M}}(C|\vec{s}_t,\operatorname{do}(\vec{a}'_t))$.
    The search process completes when $\vec{a}^*_t = \vec{a}'_t$ is found that satisfies Eq. \ref{eq:actorObjective}. A detailed description of the search procedure, including parameter settings and implementation details, can be found in~\cite{diehl2026causalapproachpredictingimproving}.
    The complete robot trajectory $\mathcal{T}$ that results from combining $\mathcal{N}$ and $\mathcal{M}$ is presumed to improve the perceived competence in the robot's navigation behaviour.
    We measure whether this is indeed the case by using the offline evaluation module from Sec. \ref{sec:methodology-critic}, which evaluates $\mathcal{T}$ after the execution is complete.

    \begin{figure}[t]
        \centering
        \includegraphics[width=\linewidth]{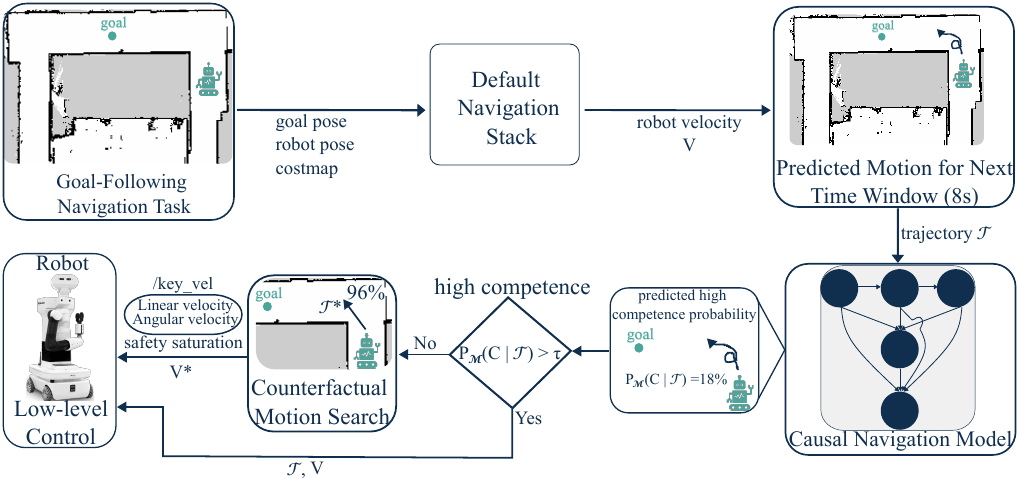}
        \caption{Online causal adaptation framework. The causal navigation model $\mathcal{M}$ estimates the competence probability for a trajectory $\mathcal{T}$ induced by the velocity commands of the default navigation stack. If  $P_{\mathcal{M}}(C|\mathcal{T}) < \text{threshold }\tau$, the online adaptation generates a counterfactual trajectory $\mathcal{T}^*$, maps $\mathcal{T}^*$ to velocity $V^*$ and sends it to the robot's controller.}
        \label{fig:causal_online_adaptation_blocks}
        \vspace{-0.3cm}
    \end{figure}

    \section{EVALUATION}
    \label{sec:evaluation}

    We evaluate the causal navigation model in an experimental study whose purpose is twofold:
    (i) we investigate whether the model can perform meaningful offline evaluation of the quality of both successful and failed autonomous navigation trajectories, and how the model's own evaluation compares with our manual evaluation, and
    (ii) we analyse whether the model can be used to perform online adaptation of the navigation behaviour of a robot.

    \subsection{Experimental Setup}

\begin{figure}[t]
        \centering
        \includegraphics[width=0.55\linewidth]{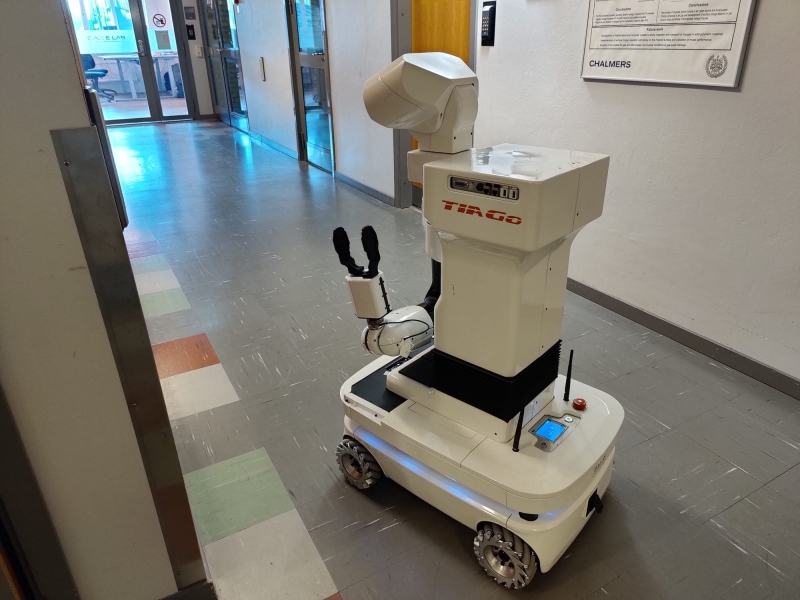}
        \caption{Our PAL TIAGo robot in a representative corridor of the experimental environment.}
        \label{fig:tiago}
        \vspace{-0.5cm}
    \end{figure}

\begin{figure}[t]
        \centering
        \includegraphics[width=0.55\linewidth]{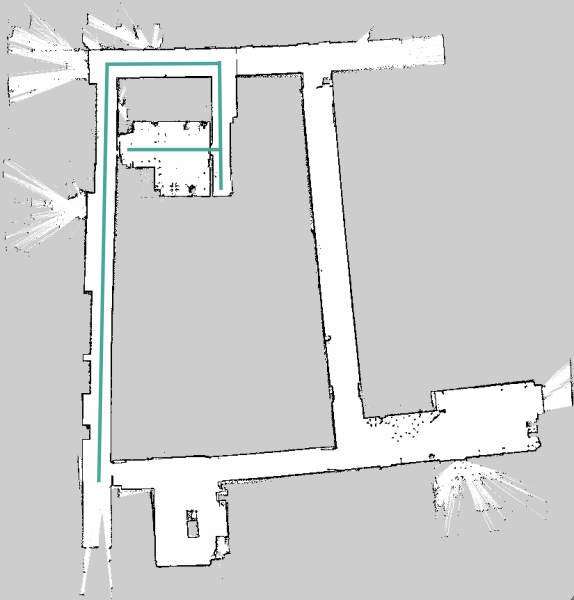}
        \caption{Navigation map used in the experiment. The region marked green is where the robot was navigating in our trials, namely, all collected trajectories are segments of this region, with varying start and end positions. The closed area where the robot is allowed to move is our robotics lab. The remaining part of the map is unused in the evaluation.}
        \label{fig:map}
        \vspace{-0.5cm}
    \end{figure}

    We perform our study with a PAL TIAGo robot, which has an omnidirectional mobile base. We only perform motions to simulate a "differential-drive" platform, which is the most common locomotion type used in these scenarios.
    The navigation includes scenarios where the robot moves both inside our research lab and in the surrounding corridors.
    Fig. \ref{fig:tiago} shows the robot in a representative corridor, while Fig. \ref{fig:map} illustrates the map used for the navigation.

    For the offline evaluation, we collected a total of $60$ goal-following navigation trajectories using the default nav2 stack, evenly divided into three categories (20 per category):
    (i) motion without obstacles (i.e., expected straight-line movement from the robot’s starting position to the goal),
    (ii) turning around a corner, and
    (iii) motion in scenarios involving humans (either passing by the robot or temporarily obstructing its motion).
    For each trajectory, the navigation goal was manually specified, and all trajectories were different in length; the majority of these were successfully performed, but a few involved the robot getting stuck during the motion and turning in place instead of rotating forward.

    For the online evaluation, we collected $30$ navigation trajectories ($10$ trajectories per category).
    To use the causal model, we created a trajectory prediction for the following $8s$, assuming constant linear and angular velocities during that period.
    By default, the velocity set by the built-in navigation component was used to drive the robot's motion, but in cases where the velocity generated by the causal model was used, we applied the causal-model velocity for $1s$ before estimating a new prediction.
    The linear and angular velocities calculated by the causal model are kept in the ranges $[-0.1m/s,0.1m/s]$ and $[-0.1rad/s,0.1rad/s]$ respectively in order to prevent too abrupt motions that might lead to collisions.
    In the calculation of $SR$, we use $\delta = 0.006\, m$ as a small tolerance to distinguish stationary behavior from actual motion, ensuring that small numerical fluctuations are not interpreted as motion.

    To ensure a fair and rigorous comparison, we established a baseline by running the robot through the exact same 30 online trajectory scenarios using only the default nav2 stack.

    \subsection{Offline Trajectory Evaluation}

    \paragraph{Correlation analysis between predicted competence and social navigation metrics} To examine whether the predicted competence variable reflects interpretable navigation behavior, we compute Spearman’s rank~\cite{hauke2011comparison} correlation $\rho$ between the model-predicted competence probability and the trajectory metrics defined in Sec.~\ref{sec:methodology}, as the correlation measures the strength of monotonic association.
    For the calculation, we discarded the windows for which a $0$ success chance was predicted, as the parameterisation is not included by the model.

\begin{figure}[t]
        \centering
        \includegraphics[width=\linewidth]{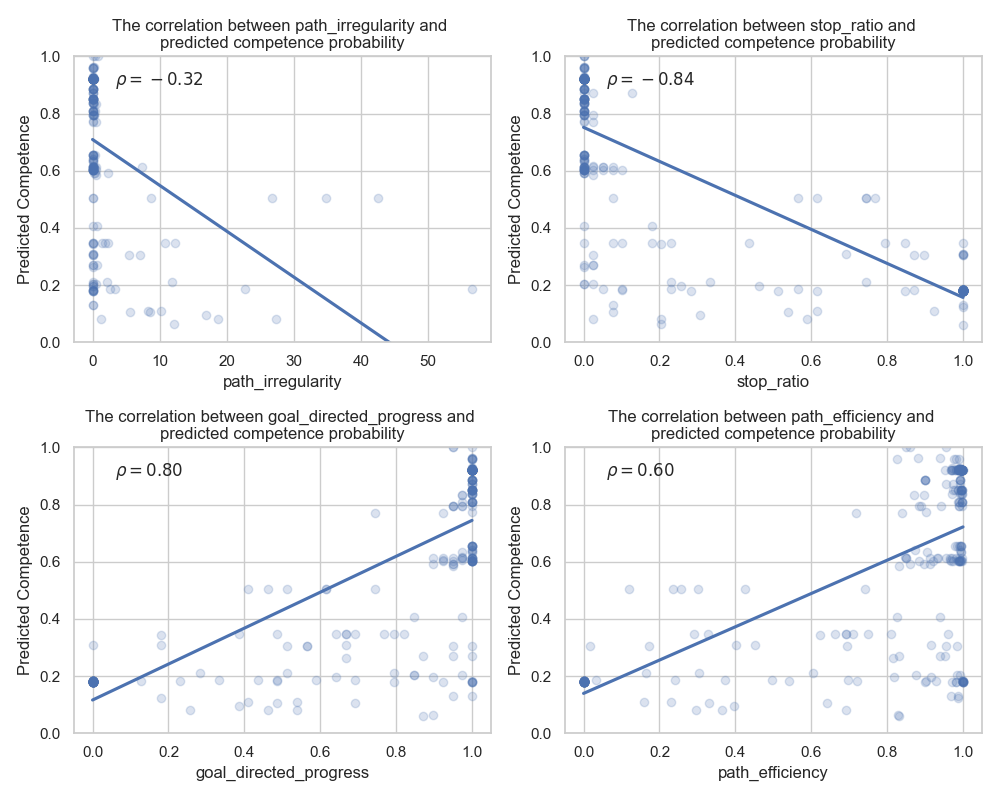}
        \caption{Correlational plots between the navigation behavior metrics and the predicted human-perceived competence.}
        \label{fig:evaluation-correlation}
        \vspace{-0.4cm}
    \end{figure}

    The results of the evaluation are shown in Tab.~\ref{tab:correlation_metrics} and Fig.~\ref{fig:evaluation-correlation}.
    As can be seen here, the metrics indicating positive behaviour (goal-directed progress and path efficiency) are both positively correlated with the competence prediction, which indicates that the model can successfully recognise the situation when the robot is making progress towards the goal.
    On the other hand, the metrics that indicate negative behaviour (path irregularity and stop ratio) are both negatively correlated with the predicted competence, which shows that the model is also able to correctly evaluate cases in which the robot's motion does not necessarily lead to goal progress.
    \paragraph{Human-model Competence Agreement} As described in Sec.~\ref{sec:methodology-actor}, we evaluate the agreement between model-predicted competence and human annotations using Cohen’s $\kappa$.
    For this, we manually annotated the competence in randomly selected $20\%$ of the trajectory windows ($57$ out of $287$).
    As shown in Tab.~\ref{tab:kappa-evaluation}, substantial to near-perfect agreement was observed for moderate thresholds (e.g., $\kappa = 0.88$ at threshold $0.6$) across all trajectories, indicating strong alignment between the probabilistic model output and human perception of navigation competence.
    When analysed by trajectory type, agreement remained high for trajectory without obstacle (i) and with obstacle (ii) scenarios at $\tau = 0.5, 0.6$, while corner trajectories (iii) exhibited a sharper performance degradation at high thresholds. This is likely because, as the complexity of the navigation scenario is increased, the robot performs larger orientation adjustments to align with the corridor and avoid obstacles. Such scenarios often involve additional turns and occasional halts to adapt motion, thereby lowering the predicted competence in the causal model. 
    These results suggest that the learned competence representation is also consistent with human annotations across diverse navigation contexts, and provide candidate values of $\tau$ in the subsequent online adaptation study.

    \begin{table}[t]
        \centering
        \caption{Spearman correlation between the predicted competence and the trajectory metrics.}
        \label{tab:correlation_metrics}
        \begin{tabular}{M{0.235\linewidth}|M{0.235\linewidth}|M{0.375\linewidth}}
            \cellcolor{gray!10!white}\textbf{Metric} & \cellcolor{gray!10!white}$\boldsymbol{\rho}$ & \cellcolor{gray!10!white}\textbf{Monotonic Trend} \\
            $GP$ & 0.80 &  Strong increasing \\
            $PE$ & 0.60 &   Moderate increasing \\
            $SR$ & -0.84 &  Strong decreasing \\
            $PI$ & -0.32 & Weak decreasing \\
        \end{tabular}
        \vspace{-0.3cm}
    \end{table}

    \begin{table}[t]
        \caption{Cohen's $\kappa$ agreement between the predicted and the annotated competence across trajectory types and probability thresholds $\tau$}
        \label{tab:kappa-evaluation}
        \centering
        \begin{tabular}{M{0.1\linewidth}|M{0.16\linewidth}| M{0.16\linewidth}|M{0.16\linewidth}|M{0.16\linewidth}}
            \cellcolor{gray!10!white}\textbf{$\boldsymbol{\tau}$} &
            \cellcolor{gray!10!white}\textbf{All} &
            \cellcolor{gray!10!white}\textbf{Straight} &
            \cellcolor{gray!10!white}\textbf{Corner} &
            \cellcolor{gray!10!white}\textbf{Obstacle} \\
            \textbf{0.5} & 0.87 & 0.74 & \textbf{0.86} & \textbf{1.00} \\
            \textbf{0.6} & \textbf{0.88} & \textbf{1.00} & \textbf{0.86} & 0.79 \\
            0.7 & 0.68 & 0.79 & 0.64 & 0.62 \\
            0.8 & 0.64 & 0.79 & 0.64 & 0.48 \\
            0.9 & 0.23 & 0.79 & 0.07 & 0.15 \\
        \end{tabular}
        \vspace{-0.3cm}
    \end{table}

    \subsection{Online Navigation Adaptation}

    \paragraph{Quantitative evaluation} In the online navigation experiment, as in the offline evaluation, we did not consider the model's predictions when the predicted success probability was $0$.
    We used a threshold $\tau = 0.5$ in all trials, as $\kappa$ was higher for the more complex corner and obstacle scenarios, and the causal intervention can be applied more frequently in those cases.
    Furthermore, we only consider trajectories in which the goal could be reached successfully (with one exception where the goal was almost reached).
    
    The average number of times the causal model was active instead of the built-in navigation component in the three scenario types is shown in Tab.~\ref{tab:causal_model_applications}.
    As the results show, the causal model did indeed occasionally interrupt the built-in component, particularly in cases where obstacles were present, where we observed that the causal model led to safer motion in general.
    \begin{table}[t]
        \centering
        \caption{Average number of causal model applications (over $10$ trajectories per scenario)}
        \label{tab:causal_model_applications}
        \begin{tabular}{M{0.275\linewidth}|M{0.275\linewidth}|M{0.275\linewidth}}
            \cellcolor{gray!10!white}\textbf{Straight} & \cellcolor{gray!10!white}\textbf{Corner} & \cellcolor{gray!10!white}\textbf{Obstacle} \\
            $4.9$ & $9.6$ & $9.3$
        \end{tabular}
        \vspace{-0.5cm}
    \end{table}

    Tab.~\ref{tab:online_navigation_performance} shows the average values of the predicted competence and the navigation metrics in the three scenarios, comparing the default navigation baseline with our causal-adapted method under identical starting and goal poses.
    In the obstacle scenario, the causal-adapted method consistently outperforms the baseline across all metrics, demonstrating that causal adaptation enhances navigation in more complex and interaction-rich environments.
    In the corner scenario, the causal method improves $GP$, $SR$, and $PI$, indicating more stable behaviour, although with a slight reduction in path efficiency.
    In the straight-line scenario, the causal method degrades performance. This is expected, as the default navigation stack already achieves near-optimal behaviour, and the causal model may introduce unnecessary adjustments (e.g, slowing down the motion) due to sensitivity to minor deviations in the predicted trajectory.
    Overall, these results indicate that causal adaptation improves performance most clearly in complex scenarios, while performing comparably to the baseline in simple navigation tasks.

    \begin{table}[t]
        \centering
        \caption{Comparison between the default nav2 baseline and the causal-adapted method across identical online scenarios (average over 10 trajectories per scenario), where both methods are executed with identical starting and goal poses.}
        \label{tab:online_navigation_performance}
        \scriptsize
        \setlength{\tabcolsep}{5pt}
        \begin{tabular}{l|cc|cc|cc}
            \cellcolor{gray!10} & \multicolumn{2}{c|}{\cellcolor{gray!10}\textbf{Straight}} & \multicolumn{2}{c|}{\cellcolor{gray!10}\textbf{Corner}} & \multicolumn{2}{c}{\cellcolor{gray!10}\textbf{Obstacle}} \\
            \cellcolor{gray!10}\textbf{Metric} 
            & \cellcolor{gray!10}\textbf{Baseline} & \cellcolor{gray!10}\textbf{Causal} 
            & \cellcolor{gray!10}\textbf{Baseline} & \cellcolor{gray!10}\textbf{Causal} 
            & \cellcolor{gray!10}\textbf{Baseline} & \cellcolor{gray!10}\textbf{Causal} \\
            \hline
            $GP$  & $\mathbf{0.992}$ & $0.991$ & $0.724$ & $\mathbf{0.940}$ & $0.870$ & $\mathbf{0.976}$ \\
            $PE$  & $\mathbf{0.981}$ & $0.979$ & $\mathbf{0.972}$ & $0.876$ & $0.869$ & $\mathbf{0.966}$ \\
            $SR$  & $\mathbf{0.012}$ & $0.057$ & $0.474$ & $\mathbf{0.107}$ & $0.118$ & $\mathbf{0.077}$ \\
            $PI$  & $\mathbf{0.0}$ & $0.021$ & $\mathbf{0.0}$ & $0.142$ & $1.414$ & $\mathbf{0.054}$ \\
            \hline
            $P_c$  & $\mathbf{92.85\%}$ & $81.48\%$ & $25.0\%$ & $\mathbf{65.22\%}$ & $74.07\%$ & $\mathbf{87.88\%}$ \\
    \end{tabular}
    \vspace{-0.3cm}
    \end{table}

    \paragraph{Qualitative observations} The results of the online experiment confirm that social causal models have the potential to be used with existing robot navigation frameworks.
    This is promising because it suggests that, rather than replacing existing frameworks, the model can be used to enhance them in more complex or dynamic scenarios.
    Nevertheless, the online experiment also demonstrated some limitations of the model.
    In some cases, the causal model-driven navigation leads to unsafe behaviour, such as motion close to or towards a wall.
    While this observation is expected, as the model does not explicitly account for environmental variables, it can be addressed by incorporating a safety filter to activate the causal adaptation module only when the costmap indicates that the surrounding environment is sufficiently clear (e.g., a filter based on a minimum free-space radius around the robot). This will be explored in future work.

    \section{DISCUSSION AND CONCLUSION}
    \label{sec:discussion}

    In this paper, we conducted a thorough evaluation of a causal navigation model, specifically one defined to improve human-perceived competence, in a variety of navigation scenarios performed by a physical robot in an everyday environment.
    Our study demonstrates that such a model can be used both as an offline evaluation module that rates the quality of offline trajectories, but also as an online adaptation module that controls the navigation behaviour of a robot.
    Additionally, the results of our evaluation confirm that competence can be a useful proxy for modelling and evaluating a robot's navigation performance.

    There are various ways in which future work can build on and improve our results.
    Firstly, it would be informative to conduct a large-scale user study in which robot navigation trajectories are evaluated by users, potentially with different robots, in order to investigate whether the robot embodiment has an effect on the perceived competence.
    Additionally, in this work, we used a causal navigation model that was manually defined for the task, based on designer insights; however, for generality, it would be useful to investigate how causal model learning can be used, such as in \cite{edstroem2023}, particularly for identifying relevant variables to include in the model.
    One major disadvantage of the causal model used in this work is that it does not encode any information about the environment in which a robot is navigating, such as static obstacles or motion patterns of dynamic obstacles, but enhancing the model to include those aspects would be essential for using the model in more complex everyday navigation scenarios.
    Alternatively, the use of different causal models could also be worthwhile to investigate, as our method is actually independent of the specific causal navigation model.
    Finally, in this study, we only performed a comparison with the default ROS2 navigation stack due to its widespread use among different robots; in future work, it would be interesting to expand the comparison to other navigation methods, such as learning- or optimization-based approaches, as well as consider additional navigation scenarios, such as navigation in a logistics center.

\addtolength{\textheight}{-4cm}

\bibliographystyle{IEEEtran}
\bibliography{references}

\begin{thebibliography}{10}
\providecommand{\url}[1]{#1}
\csname url@samestyle\endcsname
\providecommand{\newblock}{\relax}
\providecommand{\bibinfo}[2]{#2}
\providecommand{\BIBentrySTDinterwordspacing}{\spaceskip=0pt\relax}
\providecommand{\BIBentryALTinterwordstretchfactor}{4}
\providecommand{\BIBentryALTinterwordspacing}{\spaceskip=\fontdimen2\font plus
\BIBentryALTinterwordstretchfactor\fontdimen3\font minus
  \fontdimen4\font\relax}
\providecommand{\BIBforeignlanguage}[2]{{%
\expandafter\ifx\csname l@#1\endcsname\relax
\typeout{** WARNING: IEEEtran.bst: No hyphenation pattern has been}%
\typeout{** loaded for the language `#1'. Using the pattern for}%
\typeout{** the default language instead.}%
\else
\language=\csname l@#1\endcsname
\fi
#2}}
\providecommand{\BIBdecl}{\relax}
\BIBdecl

\bibitem{uhde2020}
C.~Uhde, N.~Berberich, K.~Ramirez-Amaro, and G.~Cheng, ``{The Robot as
  Scientist: Using Mental Simulation to Test Causal Hypotheses Extracted from
  Human Activities in Virtual Reality},'' in \emph{Proc. IEEE/RSJ Int. Conf.
  Intelligent Robots and Systems (IROS)}, 2020, pp. 8081--8086.

\bibitem{hellstroem2021}
T.~H. Hellstr{\"o}m, ``{The relevance of causation in robotics: A review,
  categorization, and analysis},'' \emph{Paladyn, Journal of Behavioral
  Robotics}, vol.~12, no.~1, pp. 238--255, Apr. 2021.

\bibitem{stocking2022}
K.~C. Stocking, A.~Gopnik, and C.~Tomlin, ``{From Robot Learning To Robot
  Understanding: Leveraging Causal Graphical Models For Robotics},'' in
  \emph{Proc. 5th Conference on Robot Learning (CoRL)}, 2022, pp. 1776--1781.

\bibitem{zhu2025}
X.~Zhu, L.~Bi, T.~Wu, C.~Zhang, and J.~Wu, ``{Causal Correction and
  Compensation Network for Robotics: Applications and Validation in Continuous
  Control},'' \emph{Applied Sciences}, vol.~15, no.~17, pp. 9628:1--23, 2025.

\bibitem{lee2023}
T.~E. Lee, S.~Vats, S.~Girdhar, and O.~Kroemer, ``{SCALE: Causal Learning and
  Discovery of Robot Manipulation Skills using Simulation},'' in \emph{Proc.
  7th Conf. Robot Learning (CoRL)}, 2023, pp. 2229--2256.

\bibitem{castri2025}
L.~Castri, G.~Beraldo, and N.~Bellotto, ``{Causality-enhanced Decision-Making
  for Autonomous Mobile Robots in Dynamic Environments},'' \emph{CoRR}, vol.
  abs/2504.11901, 2025.

\bibitem{fang2026}
L.~Fang, W.~Tang, and J.~Liu, ``{Counterfactual-Based Root Cause Analysis for
  Misconfigurations in Autonomous Driving Systems},'' \emph{IEEE Robotics and
  Automation Letters (RA-L)}, vol.~11, no.~1, pp. 786--793, 2026.

\bibitem{parker2006}
L.~E. Parker and B.~Kannan, ``{Adaptive Causal Models for Fault Diagnosis and
  Recovery in Multi-Robot Teams},'' in \emph{Proc. IEEE/RSJ Int. Conf.
  Intelligent Robots and Systems (IROS)}, 2006, pp. 2703--2710.

\bibitem{erdem2012}
E.~Erdem, K.~Haspalamutgil, V.~Patoglu, and T.~Uras, ``{Causality-based
  planning and diagnostic reasoning for cognitive factories},'' in \emph{Proc.
  IEEE 17th Int. Conf. Emerging Technologies \& Factory Automation (ETFA)},
  2012, pp. 1--8.

\bibitem{mitrevski2023}
A.~Mitrevski, P.~G. Pl{\"o}ger, and G.~Lakemeyer, ``{A Hybrid Skill
  Parameterisation Model Combining Symbolic and Subsymbolic Elements for
  Introspective Robots},'' \emph{Robotics and Auton. Systems}, vol. 161, pp.
  104\,350:1--22, Mar. 2023.

\bibitem{diehl2023}
M.~Diehl and K.~Ramirez-Amaro, ``{A causal-based approach to explain, predict
  and prevent failures in robotic tasks},'' \emph{Robotics and Auton. Systems},
  vol. 162, pp. 104\,376:1--12, Apr. 2023.

\bibitem{han2023}
Z.~Han and H.~Yanco, ``{Communicating Missing Causal Information to Explain a
  Robot’s Past Behavior},'' \emph{ACM Transactions on Human-Robot
  Interaction}, vol.~12, no.~1, Mar. 2023.

\bibitem{castri2022}
L.~Castri, S.~Mghames, M.~Hanheide, and N.~Bellotto, ``{Causal Discovery of
  Dynamic Models for Predicting Human Spatial Interactions},'' in \emph{Proc.
  Int. Conf. Social Robotics (ICSR)}, 2022, pp. 154--164.

\bibitem{briglia2025}
G.~Briglia, S.~Mariani, and F.~Zambonelli, ``{Towards Safe Action Policies in
  Multi-robot Systems with Causal Reinforcement Learning},'' in \emph{Workshop
  Agents and Robots for Reliable Engineered Autonomy (AREA)}, 2025, pp. 51--71.

\bibitem{wang2026}
Y.~Wang, B.~Liu, L.~Zhang, and Y.~Liu, ``{Efficient Robot Navigation in Dense
  Crowds via Causal Reasoning on Spatio-Temporal Graphs},'' \emph{Research
  Square (Preprint)}, 2026.

\bibitem{parvaneh2020}
A.~Parvaneh, E.~Abbasnejad, D.~Teney, J.~Q. Shi, and A.~van~den Hengel,
  ``{Counterfactual Vision-and-Language Navigation: Unravelling the Unseen},''
  in \emph{Advances in Neural Information Processing Systems (NeurIPS)},
  vol.~33, 2020, pp. 5296--5307.

\bibitem{vorbach2021}
C.~Vorbach \emph{et~al.}, ``{Causal Navigation by Continuous-time Neural
  Networks},'' in \emph{Advances in Neural Information Processing Systems
  (NeurIPS)}, vol.~34, 2021, pp. 12\,425--12\,440.

\bibitem{wang2023}
X.~Wang, Y.~Liu, X.~Song, B.~Wang, and S.~Jiang, ``{CaMP: Causal Multi-policy
  Planning for Interactive Navigation in Multi-room Scenes},'' in
  \emph{Advances in Neural Information Processing Systems (NeurIPS)}, vol.~36,
  2023, pp. 15\,855--15\,868.

\bibitem{zhang2023}
S.~Zhang \emph{et~al.}, ``{Layout-Based Causal Inference for Object
  Navigation},'' in \emph{Proc. IEEE/CVF Conf. Computer Vision and Pattern
  Recognition (CVPR)}, 2023, pp. 10\,792--10\,802.

\bibitem{wang2024_neur}
R.~Wang \emph{et~al.}, ``{Causality-aware transformer networks for robotic
  navigation},'' in \emph{Proc. Int. Conf. Neural Information Processing},
  2024, pp. 403--418.

\bibitem{wang2024_cvpr}
L.~Wang \emph{et~al.}, ``{Vision-and-Language Navigation via Causal
  Learning},'' in \emph{Proc. IEEE/CVF Conf. Computer Vision and Pattern
  Recognition (CVPR)}, 2024, pp. 13\,139--13\,150.

\bibitem{zhao2024}
X.~Zhao \emph{et~al.}, ``{Target-Driven Visual Navigation by Using Causal
  Intervention},'' \emph{{IEEE Transactions on Intelligent Vehicles}}, vol.~9,
  no.~1, pp. 1294--1304, 2024.

\bibitem{liu2025}
R.~Liu, S.~Wu, D.~Lin, and W.~Zhang, ``{CVLN-Think: Causal Inference with
  Counterfactual Style Adaptation for Continuous Vision-and-Language
  Navigation},'' in \emph{Proc. IEEE/RSJ Int. Conf. Intelligent Robots and
  Systems (IROS)}, 2025, pp. 15\,299--15\,305.

\bibitem{wang2025}
Z.~Wang, H.~Hu, W.~Gao, and S.~Shen, ``{Causal Enhanced Autoregressive Model
  for Monocular Image-Goal Navigation in Unknown Map Environment},'' \emph{IEEE
  Robotics and Automation Letters (RA-L)}, vol.~10, no.~8, pp. 8364--8371,
  2025.

\bibitem{cannizzaro2025}
R.~Cannizzaro \emph{et~al.}, ``{COBRA-PPM: A Causal Bayesian Reasoning
  Architecture Using Probabilistic Programming for Robot Manipulation Under
  Uncertainty},'' in \emph{Proc. European Conf. Mobile Robots (ECMR)}, 2025,
  pp. 1--8.

\bibitem{tung2024}
Y.-S. Tung, H.~Gupta, W.~Jiang, B.~Hayes, and A.~Roncone, ``{Causal Influence
  Detection for Human Robot Interaction},'' in \emph{HRI Workshop on Causal
  Learning for Human-Robot Interaction}, 2024.

\bibitem{mcduff2022}
D.~McDuff \emph{et~al.}, ``{CausalCity: Complex Simulations with Agency for
  Causal Discovery and Reasoning},'' in \emph{Proc. 1st Conf. Causal Learning
  and Reasoning}, vol. 177, 2022, pp. 559--575.

\bibitem{gao2022evaluation}
Y.~Gao and C.-M. Huang, ``{Evaluation of socially-aware robot navigation},''
  \emph{Frontiers in Robotics and AI}, vol.~8, p. 721317, 2022.

\bibitem{mavrogiannis2022social}
C.~Mavrogiannis, P.~Alves-Oliveira, W.~Thomason, and R.~A. Knepper, ``{Social
  momentum: Design and evaluation of a framework for socially competent robot
  navigation},'' \emph{ACM Trans. Human-Robot Interaction (THRI)}, vol.~11,
  no.~2, pp. 1--37, 2022.

\bibitem{tsoi2022sean}
N.~Tsoi, A.~Xiang, P.~Yu, S.~S. Sohn, G.~Schwartz, S.~Ramesh, M.~Hussein, A.~W.
  Gupta, M.~Kapadia, and M.~V{\'a}zquez, ``Sean 2.0: Formalizing and generating
  social situations for robot navigation,'' \emph{IEEE Robotics and Automation
  Letters}, vol.~7, no.~4, pp. 11\,047--11\,054, 2022.

\bibitem{zhang2025predicting}
Q.~Zhang, N.~Tsoi, M.~Nagib, B.~Choi, J.~Tan, H.-T.~L. Chiang, and
  M.~V{\'a}zquez, ``Predicting human perceptions of robot performance during
  navigation tasks,'' \emph{ACM Transactions on Human-Robot Interaction},
  vol.~14, no.~3, pp. 1--27, 2025.

\bibitem{zhang2025few}
Q.~Zhang, N.~Tsoi, M.~Nagib, H.-T.~L. Chiang, and M.~V{\'a}zquez, ``Few-shot
  inference of human perceptions of robot performance in social navigation
  scenarios,'' \emph{arXiv preprint arXiv:2512.16019}, 2025.

\bibitem{hossen2023}
M.~A. Hossen \emph{et~al.}, ``{CaRE: Finding Root Causes of Configuration
  Issues in Highly-Configurable Robots},'' \emph{IEEE Robotics and Automation
  Letters (RA-L)}, vol.~8, no.~7, pp. 4115--4122, 2023.

\bibitem{pearl1994}
J.~Pearl, ``{A Probabilistic Calculus of Actions},'' in \emph{Proc. 10th Conf.
  Uncertainty in Artificial Intelligence}, 1994, pp. 454--462.

\bibitem{JSSv035i03}
\BIBentryALTinterwordspacing
M.~Scutari, ``Learning bayesian networks with the bnlearn r package,''
  \emph{Journal of Statistical Software}, vol.~35, no.~3, p. 1–22, 2010.
  [Online]. Available:
  \url{https://www.jstatsoft.org/index.php/jss/article/view/v035i03}
\BIBentrySTDinterwordspacing

\bibitem{diehl2026causalapproachpredictingimproving}
\BIBentryALTinterwordspacing
M.~Diehl, N.~Tsoi, G.~Chavez, K.~Ramirez-Amaro, and M.~Vázquez, ``A causal
  approach to predicting and improving human perceptions of social navigation
  robots,'' 2026. [Online]. Available: \url{https://arxiv.org/abs/2603.11290}
\BIBentrySTDinterwordspacing

\bibitem{biswas2022socnavbench}
A.~Biswas \emph{et~al.}, ``{Socnavbench: A grounded simulation testing
  framework for evaluating social navigation},'' \emph{ACM Trans. Human-Robot
  Interaction (THRI)}, vol.~11, no.~3, pp. 1--24, 2022.

\bibitem{schneider2005discussion}
F.~E. Schneider, D.~Wildermuth, and A.~Kr{\"a}u{\ss}ling, ``Discussion of
  exemplary metrics for multi-robot systems for formation navigation,''
  \emph{International Journal of Advanced Robotic Systems}, vol.~2, no.~4,
  p.~37, 2005.

\bibitem{cohen1960}
J.~Cohen, ``{A coefficient of agreement for nominal scales},''
  \emph{Educational and psychological measurement}, vol.~20, no.~1, pp. 37--46,
  1960.

\bibitem{hauke2011comparison}
J.~Hauke and T.~Kossowski, ``Comparison of values of pearson's and spearman's
  correlation coefficients on the same sets of data,'' \emph{Quaestiones
  geographicae}, vol.~30, no.~2, pp. 87--93, 2011.

\bibitem{edstroem2023}
F.~Edstr{\"o}m, T.~Hellstr{\"o}m, and X.~De~Luna, ``{Robot Causal Discovery
  Aided by Human Interaction},'' in \emph{Proc. 32nd IEEE Int. Conf. Robot and
  Human Interactive Communication (RO-MAN)}, 2023, pp. 1731--1736.

\end{thebibliography}

\end{document}